\documentclass[a4paper,10pt,twocolumn]{article}
\usepackage[utf8]{inputenc}
\usepackage{multicol,graphicx}
\usepackage{mathptmx}
\usepackage[T1]{fontenc}
\usepackage{textcomp}
\usepackage{url}

\usepackage{booktabs}
\usepackage{amssymb}
\usepackage{multirow}
\usepackage[T1]{fontenc}
\usepackage[english]{babel}

\usepackage[backend=bibtex,style=ieee,doi=false,isbn=false,url=true,maxnames=6,citestyle=numeric-comp,giveninits=true]{biblatex}

\fontfamily{ptm}\selectfont

\bibliography{references.bib}

\usepackage[font=small]{caption}
\usepackage{amsmath}

\newcommand{\mysection}[1]{\vspace{0.4cm} \uppercase{#1} \vspace{0.4cm}}
\newcommand{\mysubsection}[1]{\hspace{10pt}\textit{#1:}}

\begin{document}
	
\setlength{\textfloatsep}{10pt plus 1.0pt minus 2.0pt}	
\setlength{\columnsep}{1cm}

\twocolumn[%
\begin{@twocolumnfalse}
\begin{center}
	{\fontsize{14}{18}\selectfont
        \textbf{{Riemannian Geometry-Preserving Variational Autoencoder for MI-BCI Data Augmentation}}\\}
    \begin{large}
        \vspace{0.6cm}
        Viktorija Poļaka\textsuperscript{1}, Ivo Pascal De Jong\textsuperscript{1}, Andreea Ioana Sburlea\textsuperscript{1}\\
        \vspace{0.6cm}
        \textsuperscript{1}Faculty of Science and Engineering, University of Groningen, Groningen, The Netherlands\\
        \vspace{0.5cm}
        E-mail: victoria\_polaka@proton.me
        \vspace{0.4cm}
    \end{large}
\end{center}	
\end{@twocolumnfalse}%
]%

ABSTRACT: %
This paper addresses the challenge of generating synthetic electroencephalogram (EEG) covariance matrices for motor imagery brain-computer interface (MI-BCI) applications. \textit{Objective}. We aim to develop a generative model capable of producing high-fidelity synthetic covariance matrices while preserving their symmetric positive-definite nature. \textit{Approach}. We propose a Riemannian geometry-preserving variational autoencoder (RGP-VAE) integrating geometric mappings with a composite loss function combining Riemannian distance, tangent space reconstruction accuracy and generative diversity. \textit{Results}. The model generates valid, representative EEG covariance matrices, while learning a subject-invariant latent space. Synthetic data proves practically useful for MI-BCI, with its impact depending on the paired classifier. \textit{Contribution}. This work introduces and validates the RGP-VAE as a geometry-preserving generative model for EEG covariance matrices, highlighting its potential for signal privacy, scalability and data augmentation.

\mysection{introduction}\label{sec:introduction} 

While Riemannian geometry-based classifiers currently dominate MI-BCI competitions, their advancement towards mainstream applications is hindered by data scarcity and inter-subject variability, which necessitates lengthy calibration sessions \cite{congedo2017riemannian, yger2017riemannian, chevallier2024moabb, blankertz2007non}. Deep learning alternatives have yet to surpass these geometric pipelines, possibly explained by the limited availability of subject-level data \cite{chevallier2024moabb}. To overcome these limitations, we propose a novel data augmentation framework tailored to the specific geometric properties of EEG covariance matrices, which are symmetric positive-definite (SPD), i.e., symmetric with strictly positive eigenvalues.

Previous work exploring data augmentation directly on the SPD manifold by geometrically interpolating between existing covariance matrices of the same class has successfully boosted BCI classification accuracy in data-scarce scenarios for SSVEP and ERP tasks \cite{kalunga:data_aug_interp}. However, this approach is fundamentally limited to the convex hull of the original data and thus cannot generate plausible variations that exist in unexplored regions of the manifold. A variational autoencoder (VAE) \cite{vaepaper}, which can learn a latent representation of a manifold \cite{riemgeoofdeepgenmodel}, may offer an alternative to overcome this convex-hull limitation and generate diverse yet plausible synthetic samples that extend beyond the convex hull.

However, standard VAEs assume Euclidean geometry, creating a conflict when working with the Riemannian SPD manifold structure of EEG covariance matrices; applying standard Euclidean operations on this curved manifold causes geometric distortions (e.g., the "swelling effect") \cite{kalunga:data_aug_interp}. We address this by proposing the Riemannian geometry-preserving VAE (RGP-VAE) designed to preserve geometric integrity, utilizing parallel transport \cite{Yair2019Parallel} to align data and thus enable the model to learn subject-invariant features. The focus is specifically on the challenging and practically relevant problem of cross-subject generalization, with the aim to reduce the need for extensive calibration \cite{congedo2017riemannian}. Accordingly, this paper aims to: (1) establish if a Riemannian geometry-preserving VAE can generate valid synthetic EEG covariance matrices, and (2) evaluate whether this synthetic data improves cross-subject MI-BCI performance.

\mysection{METHODS}\label{sec:Methodology}
\begin{figure*}[h]
    \centering
    \includegraphics[width=360pt]{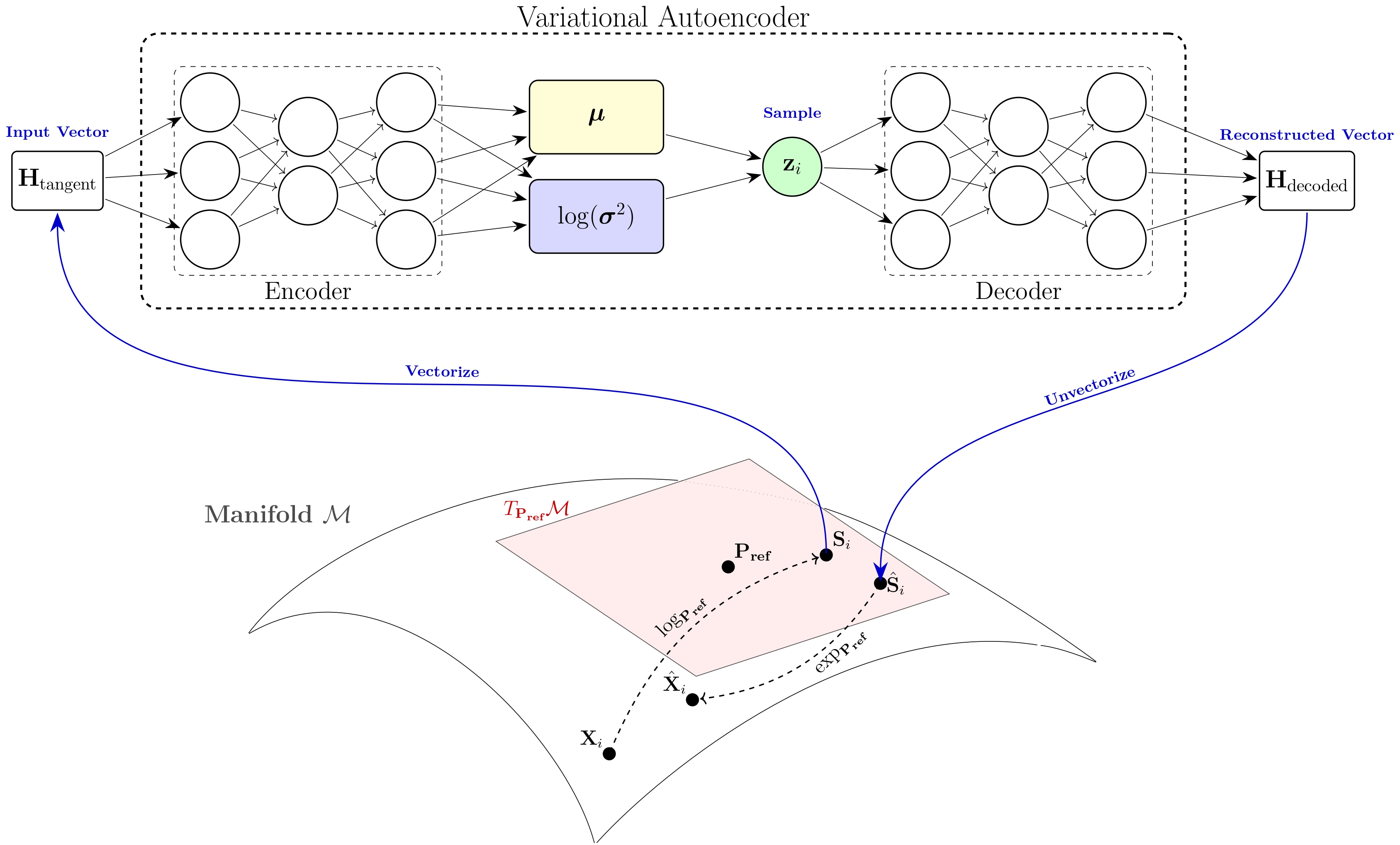}
    \caption{An overview of the proposed RGP-VAE, illustrating the integration of a standard VAE with geometric operations on the SPD manifold. An input SPD matrix $\mathbf{X}_i$ is first projected onto the tangent space at a reference point $\mathbf{P}_{\text{ref}}$ using the logarithmic map $\text{log}_{\mathbf{P}_{\text{ref}}}$ (Eq. \ref{eq:log_map}). This tangent representation $\mathbf{S}_i$ is then vectorized to serve as the encoder input $\mathbf{H}_{\text{tangent}}$. The encoder maps this input to a latent distribution parameterized by $\boldsymbol{\mu}$ and $\text{log}(\boldsymbol{\sigma}^2)$, from which a latent vector $\mathbf{z}_i$ is sampled and passed to the decoder to produce the reconstructed vector $\mathbf{H}_{\text{decoded}}$. The vector is unvectorized back into a tangent space representation $\hat{\mathbf{S}}_i$, which is finally mapped back onto the SPD manifold via the exponential map ($\text{exp}_{\mathbf{P}_{\text{ref}}}$) (Eq. \ref{eq:exp_map}) to produce the reconstructed SPD matrix $\hat{\mathbf{X}}_i$.}
    \label{fig:architecture}
\end{figure*}

\mysubsection{Data and Preprocessing}\label{sec:Preprocessing}\label{sec:datasets} 
We use the dataset from Faller et al. \cite{BNCI2015_001}, containing 13-channel EEG recordings from 12 subjects performing a two-class motor imagery task (right hand versus both feet). The data loading procedure, using the "Mother of All BCI Benchmarks" framework \cite{Aristimunha_Mother_of_all_2023}, resulted in a total of 5572 trials across the 12 subjects (398 or 597 trials per individual). %

The EEG trials are bandpass filtered (8--30 Hz) to capture sensorimotor rhythms. The raw voltage signals were then scaled to microvolts ($10^{6}$). To address non-stationarity, exponential moving standardization (EMS) \cite{SchirrmeisterSF17} is applied. EMS can be seen applied to data for training deep learning models as these models tend to be sensitive to the input scale \cite{EMS}. Finally, trials are converted into spatial covariance matrices in $\mathbb{R}^{13 \times 13}$ using the oracle approximating shrinkage estimator \cite{Chen_2010}, yielding  well-conditioned SPD matrices.

To address the inherent variability between individuals' EEG signal characteristics, which manifests as geometric differences in their location on the Riemannian manifold, parallel transport \cite{Yair2019Parallel} was applied. This technique geometrically transports matrices from each subject-specific reference mean to a global (class) reference mean via a congruence transformation.

\mysubsection{Model Architecture}
Conceptually building on prior work on Riemannian variational autoencoders for manifold-valued data \cite{Miolane2019LearningWS}, the modified VAE (Fig. \ref{fig:architecture}) learns a latent representation $\mathbf{z}$ from SPD matrices by bridging the curved manifold $\mathcal{M}$ and the Euclidean space required by neural networks. The manifold of symmetric positive-definite matrices is defined as $\mathcal{M} = \{\mathbf{X} \in \mathbb{R}^{N \times N} \mid \mathbf{X} = \mathbf{X}^\top, \mathbf{X} \succ 0\}$, where ${N}$ is the number of EEG channels and $X \succ 0$ indicates that the matrix is composed of strictly positive eigenvalues \cite{yger2017riemannian}. The proposed architecture relies on a class-specific reference point $\mathbf{P}_{\text{ref}}$, calculated as the Riemannian Fréchet mean of the training class. Unlike arithmetic mean, which may not yield a valid SPD matrix, the Fréchet Mean guarantees a valid point on the manifold by minimizing the sum of squared Riemannian distances to all other matrices in a set $\{\mathbf{X}_i\}_{i=1}^{M_{\text{set}}}$:
\begin{equation}
\mathbf{G} = \underset{\mathbf{P} \in \mathcal{M}}{\arg\min} \sum_{i=1}^{M_{set}} d_r^2(\mathbf{P}, \mathbf{X}_i)
\label{eq:riemannian_mean}
\end{equation}
where $\mathbf{P}$ is the candidate SPD matrix over which the minimization occurs and $d_r(\mathbf{P}, \mathbf{X}_i)$ is the affine-invariant Riemannian metric (AIRM)\cite{kalunga:data_aug_interp, Fletcher2004, Moakher2005} which defines the distance between two SPD matrices as:
\begin{equation}
d_r(\mathbf{P}_1, \mathbf{P}_2) = \|\log(\mathbf{P}_1^{-1/2} \mathbf{P}_2 \mathbf{P}_1^{-1/2})\|_F
\label{eq:airm}
\end{equation}

The model processes a batch of aligned SPD covariance matrices
$\mathcal{X}=\{\mathbf{X}_1,\ldots,\mathbf{X}_B\}$, with batch size $B = 128$, selected as a balance between computational load and the requirements for the diversity loss.
Each $\mathbf{X}_i$ is projected to the tangent space---a local Euclidean approximation---at the class-specific reference point $\mathbf{P}_{\mathrm{ref}}$
using the logarithmic map \cite{yger2017riemannian, congedo2017riemannian}:
\begin{equation}
\mathbf{S}_i=\log_{\mathbf{P}_{\mathrm{ref}}}(\mathbf{X}_i)
=\mathbf{P}_{\mathrm{ref}}^{1/2}\,\log\!\Big(\mathbf{P}_{\mathrm{ref}}^{-1/2}\mathbf{X}_i\mathbf{P}_{\mathrm{ref}}^{-1/2}\Big)\,\mathbf{P}_{\mathrm{ref}}^{1/2}.
\label{eq:log_map}
\end{equation}
This is implemented via batched whitening followed by the matrix logarithm to support numerical stability by centring operations around the identity.%

The resulting batch $\mathcal{S}=\{\mathbf{S}_1,\ldots,\mathbf{S}_B\}$ consists of symmetric matrices in the tangent space at $\mathbf{P}_{\mathrm{ref}}$ and each $\mathbf{S}_i$ is vectorized by using only the upper-triangular elements forming the batch of vectors $\mathbf{H}_{\mathrm{tangent}}\in\mathbb{R}^{B\times D_{\mathrm{spd}}}$ (with $D_{\mathrm{spd}}=N(N+1)/2$) as input to the encoder.

The encoder maps this batch of vectors to the parameters ($\mathbf{M}, \log\mathbf{\sigma}^2$) of the latent distribution, where $\mathbf{M} = \{\boldsymbol{\mu}_1, \ldots, \boldsymbol{\mu}_B\}$ and $\log\mathbf{\sigma}^2 = \{\log\boldsymbol{\sigma}_1^2, \ldots, \log\boldsymbol{\sigma}_B^2\}$. The encoder consists of five sequential blocks (linear → batch normalization → LeakyReLU \cite{maas2013rectifier}) with dimensions $D_{\text{spd}} \to 32 \to 64 \to 16 \to 32 \to 64$, followed by two separate linear projections to produce $\boldsymbol{\mu}_i, \log\boldsymbol{\sigma}^2_i \in \mathbb{R}^{D_{\text{lat}}}$ where $D_{\text{lat}} = 64$. Batch normalization stabilizes training by reducing internal covariate shift \cite{batch_norm}, while LeakyReLU activations are used to preserve the network's representational capacity for tangent space vectors preventing permanently inactive neurons \cite{maas2013rectifier}. The batch of latent vectors $\mathbf{Z} = \{\mathbf{z}_1, \ldots, \mathbf{z}_B\} \in \mathbb{R}^{B \times D_{\text{lat}}}$ is sampled via the reparameterization trick: $\mathbf{z}_i = \boldsymbol{\mu}_i + \boldsymbol{\epsilon}_i \odot \exp(0.5 \cdot \log\boldsymbol{\sigma}^2_i)$, where $\boldsymbol{\epsilon}_i \sim \mathcal{N}(\mathbf{0}, \mathbf{I})$, allowing gradients to flow back through $\mathbf{M}$ and $\log\boldsymbol{\Sigma}^2$ during training.

The decoder MLP mirrors the encoder structure, mapping the batch of latent vectors $\mathbf{Z}$ back to a batch of decoded vectors $\mathbf{H}_{\text{decoded}} \in \mathbb{R}^{B \times D_{\text{spd}}}$, which is subsequently unvectorized into a batch of symmetric matrices $\hat{\mathbf{S}}' \in \mathbb{R}^{B \times N \times N}$. The decoder output $\hat{\mathbf{S}}'_i$ is explicitly re-symmetrized via $\hat{\mathbf{S}}''_i = (\hat{\mathbf{S}}'_i + (\hat{\mathbf{S}}'_i)^T)/2$ to eliminate any asymmetries. To return to the manifold, we apply the Exponential Map to each matrix:
\begin{equation}
    \hat{\mathbf{X}}_i = \text{exp}_{\mathbf{P}_{\text{ref}}}(\hat{\mathbf{S}}'_{i}) = \mathbf{P}_{\text{ref}}^{1/2} \text{exp}\left(\mathbf{P}_{\text{ref}}^{-1/2} \hat{\mathbf{S}}'_{i} \mathbf{P}_{\text{ref}}^{-1/2}\right) \mathbf{P}_{\text{ref}}^{1/2}
    \label{eq:exp_map}
\end{equation}
Numerical instability caused by floating-point arithmetic can violate the strict SPD constraints, therefore validity is enforced throughout the model architecture and parallel transport. During the matrix exponential computation, eigenvalues are conditionally scaled (threshold $T=20$) to prevent overflow: if $\lambda_{\max} > T$, all eigenvalues are scaled by $T/\lambda_{\max}$. Throughout all geometric operations, we maintain a numerical threshold $\epsilon = 10^{-6}$. If the minimum eigenvalue $\lambda_{\min}$ of any intermediate or output matrix falls below $\epsilon$, we add $(\epsilon - \lambda_{\min})\mathbf{I}$ to shift all eigenvalues above the threshold, ensuring positive-definiteness.

\mysubsection{Training and Optimization}
The network is optimized using a loss function $L_{\text{total}}$ balancing reconstruction accuracy, latent space regularization, and diversity:
\begin{equation}
L_{\text{total}} = (L_{\text{manifold}} + L_{\text{tangent}}) + \beta L_{\text{KL}} + \gamma L_{\text{diversity}}
\end{equation}
The reconstruction term combines $L_{\text{manifold}}$ which enforces geometric fidelity using the AIRM distance (Eq.~\ref{eq:airm}):
\begin{equation}
L_{\text{manifold}} = \frac{1}{B} \sum_{i=1}^{B} d_r(\mathbf{X}_i, \hat{\mathbf{X}}_i)
\end{equation}
and $L_{\text{tangent}}$, which minimized the normalized Euclidean error between original and decoded tangent vectors:
\begin{equation}
L_{\text{tangent}} = \frac{1}{B}\sum_{i=1}^{B} \frac{\sum_j (h_{\text{decoded},i,j} - h_{\text{tangent},i,j})^2}{\sum_j h^2_{\text{tangent},i,j} + \epsilon}
\end{equation}
where $\epsilon = 10^{-6}$ for numerical stability. Meanwhile, latent space regularization is achieved through KL divergence $L_{\text{KL}}$ toward a standard Gaussian prior:
\begin{equation}
L_{\text{KL}} = \frac{1}{B}\sum_{i=1}^{B} \left[-0.5 \sum_{k=1}^{D_{\text{lat}}} (1 + \log\sigma^2_{i,k} - \mu^2_{i,k} - \exp(\log\sigma^2_{i,k}))\right]
\end{equation}
We apply KL cost annealing \cite{annealingbowman}, linearly increasing $\beta$ from 0.0001 to 0.2 during training to prevent posterior collapse while maintaining reconstruction fidelity.

The diversity loss $L_{\text{diversity}}$ encourages sample diversity by maximizing the geometric volume of generated tangent vectors. Since the determinant of a covariance matrix quantifies the generalized variance (i.e., the volume spanned by data points), maximizing it promotes wider spatial coverage in the tangent space. The loss minimizes the negative log-determinant of the batch covariance of decoded tangent space vectors $\mathbf{H}_{\text{decoded}} \in \mathbb{R}^{B \times D_{\text{spd}}}$:
\begin{equation}
L_{\text{diversity}} = -\log\det(\text{Cov}(\mathbf{H}^T_{\text{decoded}}) + \epsilon_{\text{cov}}\mathbf{I})
\end{equation}
with $\epsilon_{\text{cov}} = 10^{-6}$ for numerical stability, weighted by an empirically determined $\gamma = 0.035$ .

The AdamW optimizer \cite{adamW} is employed for a fixed 100 epochs with an empirically found learning rate of $1 \times 10^{-4}$ and a weight decay parameter of $1 \times 10^{-6}$. Training is further regularized with gradient clipping (max norm=1.0) and learning rate reduction (factor=0.5) after 20 epochs of stagnation. 

\mysubsection{Data Generation and Evaluation Protocol}
leave-one-subject-out cross-validation (LOSO-CV) is employed; in each fold, class-specific RGP-VAEs are trained on aligned data from $N-1$ subjects to test generalization to unseen individuals. Two synthetic generation strategies are evaluated: Posterior sampling encodes each training matrix $\mathbf{X}_i$, samples $\mathbf{z}_i$ via reparameterization, and decodes to create variations preserving core characteristics of each sample (1:5 real-to-synthetic ratio). Prior sampling draws $\mathbf{z} \sim \mathcal{N}(\mathbf{0}, \mathbf{I})$ directly to generate novel samples beyond the training convex hull (5000 per class).

Three classifiers---minimum distance to mean (MDM), k-nearest neighbors (KNN), and support vector classifier (SVC)---are trained and evaluated on held-out test subjects under three conditions: (1) baseline using only original training data, (2) augmented with synthetic data, and (3) synthetic-only training to assess standalone quality. Balanced accuracy, averaged across all folds, serves as the primary metric due to its robustness against class imbalances from potential artifact removal.

Synthetic data quality is assessed by verifying SPD properties (symmetry and positive-definiteness), comparing statistical variance (element-wise and global) between real and synthetic matrices, and measuring geometric spread via mean pair-wise Riemannian distances within each class. A scrambled-label diagnostic test confirms that performance degrades to chance level, indicating no spurious correlations. \\

Source code and additional information can be found at \url{https://641e16.github.io/RGP-VAE/}.

\begin{figure}[t]
    \centering
    \includegraphics[width=\linewidth]{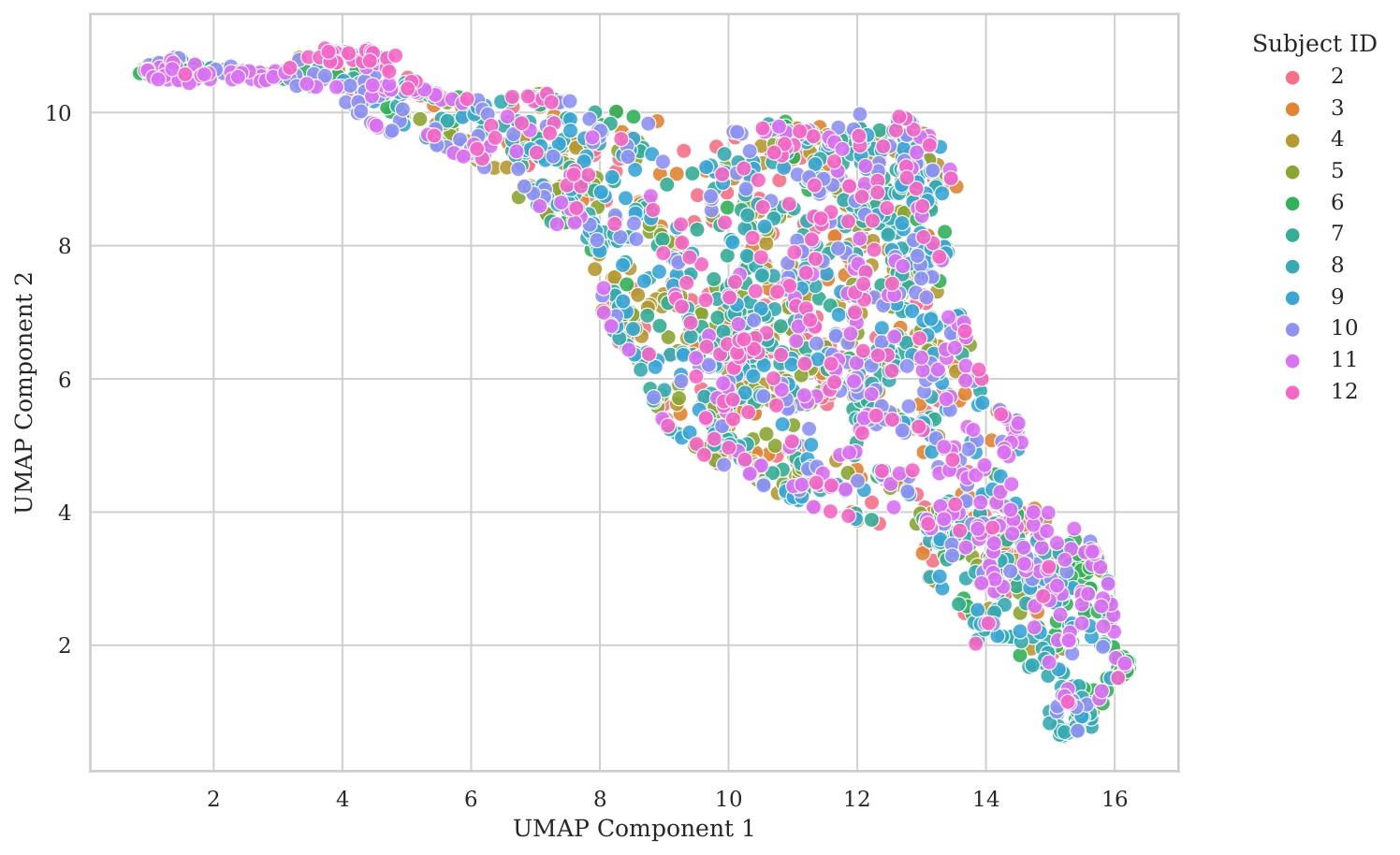}
    \caption{2D UMAP visualization of the latent space of the RGP-VAE for right-hand movement data. Points are colored by Subject ID; their significant overlap indicates the learning of a subject-invariant representation.}
    \label{fig:umap_space}
\end{figure}

\begin{figure*}[t]
    \centering
    \includegraphics[width=\textwidth]{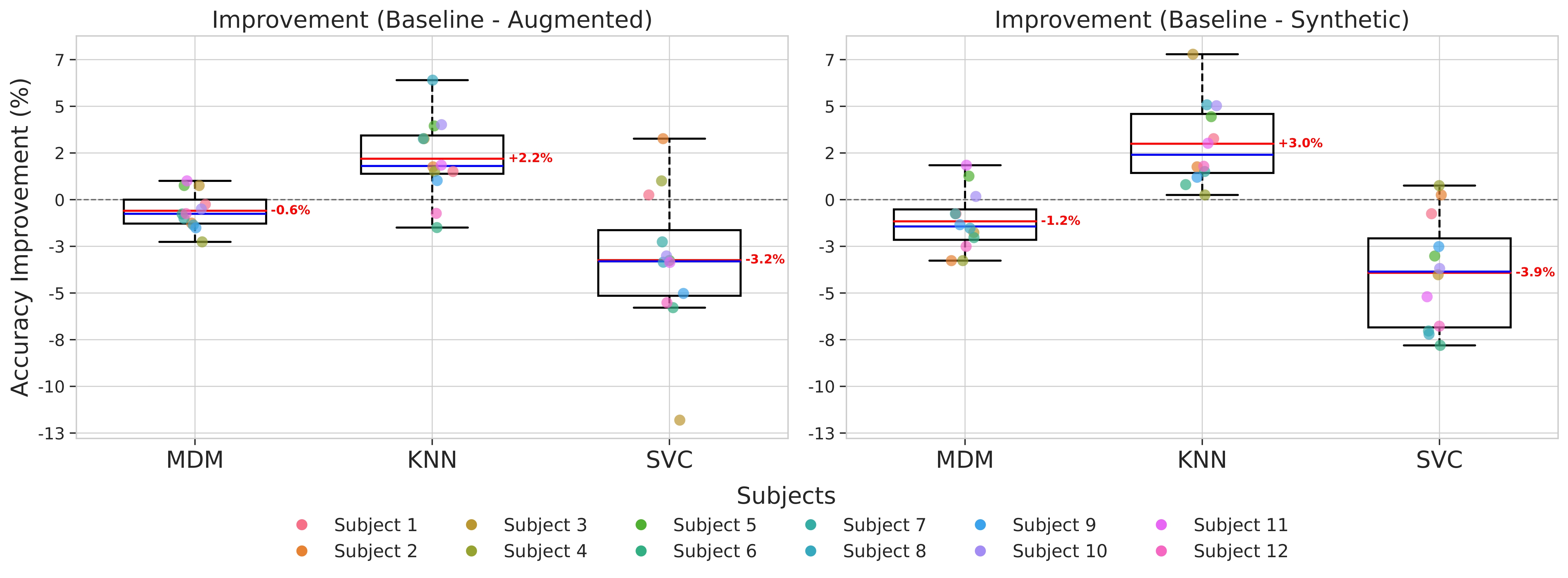}
    \caption{Distribution of accuracy improvement for each classifier using the prior generator. The plot shows the percentage point difference between the
    `Augmented' and `Synthetic-Only' conditions relative to the `Baseline' across all subjects. The red line signifies the mean whilst the blue line is the median.}
    \label{fig:improvement_dist_prior}
\end{figure*}

\begin{figure*}[t]
    \centering
    \includegraphics[width=\textwidth]{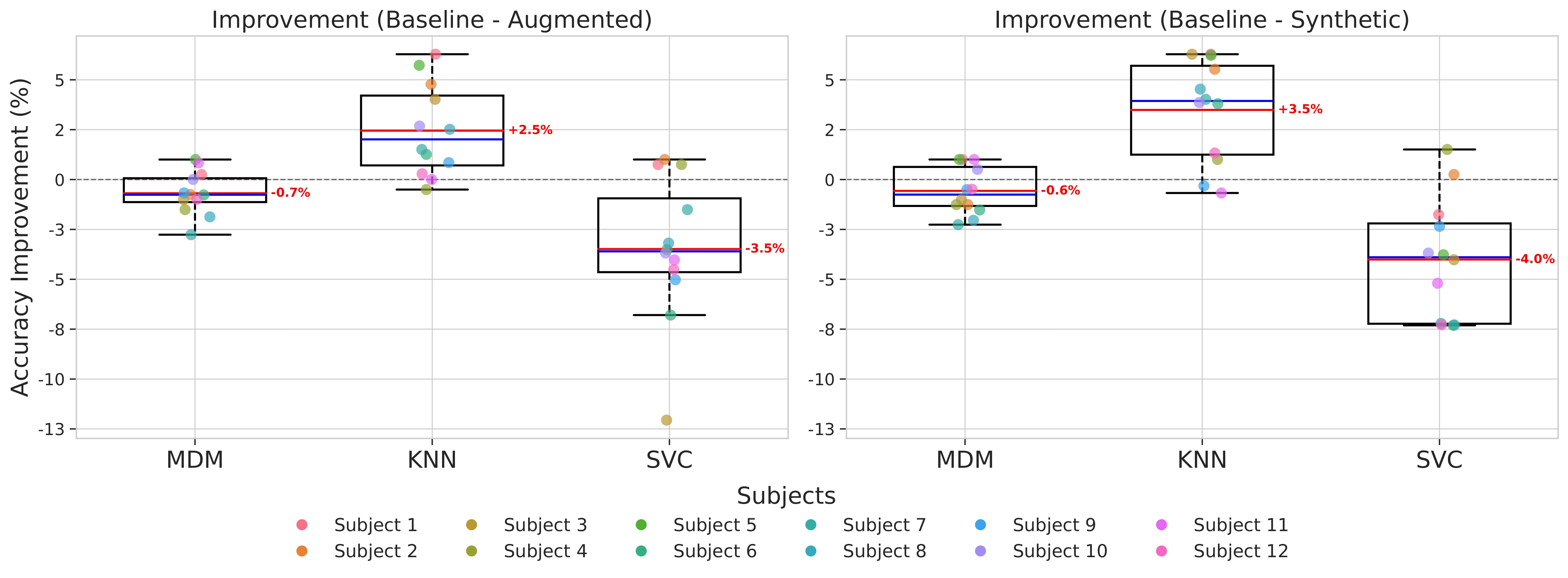}
    \caption{Distribution of accuracy improvement for each classifier using the posterior generator, showing similar trends to the prior generator but with more pronounced fluctuations.}
    \label{fig:improvement_dist_posterior}
\end{figure*}

\begin{table*}[t]
\centering
\caption{Fidelity analysis of synthetic data averaged across 12 folds. The table compares the statistical variance ratio and the mean intra-class Riemannian distance, showing that the synthetic data distribution is valid.}
\label{tab:variance_analysis}
\begin{tabular}{lcccc}
\toprule
\textbf{Generator} & \multicolumn{2}{c}{\textbf{Statistical Variance}} & \multicolumn{2}{c}{\textbf{Geometric Diversity}} \\
\cmidrule(lr){2-3} \cmidrule(lr){4-5}
& \textbf{Original} & \textbf{Synthetic (Ratio)} & \textbf{Original} & \textbf{Synthetic} \\ 
\midrule
Prior     & 0.208 & 0.221 (1.061) & 2.032 & 1.946 \\
Posterior & 0.208 & 0.221 (1.063) & 2.032 & 1.918 \\ 
\bottomrule
\end{tabular}
\end{table*}

\begin{table*}[t]
\centering
\caption{Average balanced accuracy ($\%$) across 12 subjects for all training conditions and generators with corresponding p-values.}
\label{tab:main_results}
\resizebox{\textwidth}{!}{
\begin{tabular}{@{}llccccccc@{}}
\toprule
\multirow{2}{*}{\textbf{Generator}} & \multirow{2}{*}{\textbf{Classifier}} & \textbf{Baseline} & \multicolumn{3}{c}{\textbf{Augmented Scenario}} & \multicolumn{3}{c}{\textbf{Synthetic-Only Scenario}} \\
\cmidrule(lr){4-6} \cmidrule(l){7-9}
& & \textbf{Acc. (\%)} & \textbf{Acc. (\%)} & \textbf{Improvement} & \textbf{p-value} & \textbf{Acc. (\%)} & \textbf{Improvement} & \textbf{p-value} \\ \midrule
\multirow{3}{*}{Prior} & MDM & $59.52 \pm 5.52$ & $58.92 \pm 5.40$ & -0.59\% & 0.092 & $58.36 \pm 5.03$ & -1.16\% & 0.043 \\
& KNN & $53.19 \pm 4.00$ & $55.38 \pm 4.17$ & +2.19\% & \textbf{0.003} & $56.19 \pm 4.19$ & +3.00\% & $\mathbf{<0.001}$ \\
& SVC & $60.67 \pm 5.33$ & $57.43 \pm 6.32$ & -3.24\% & 0.016 & $56.75 \pm 6.37$ & -3.92\% & \textbf{0.002} \\ \midrule
\multirow{3}{*}{Posterior} & MDM & $59.52 \pm 5.52$ & $58.83 \pm 5.29$ & -0.69\% & 0.092 & $58.95 \pm 5.51$ & -0.57\% & 0.151 \\
& KNN & $53.19 \pm 4.00$ & $55.64 \pm 4.13$ & +2.45\% & \textbf{0.002} & $56.68 \pm 4.06$ & +3.49\% & \textbf{0.002} \\
& SVC & $60.67 \pm 5.33$ & $57.18 \pm 6.57$ & -3.48\% & \textbf{0.007} & $56.66 \pm 6.25$ & -4.01\% & \textbf{0.002} \\ \bottomrule
\end{tabular}%
}
\end{table*}
\mysection{results}\label{sec:results}

We validate the proposed RGP-VAE through an assessment of the generated synthetic data fidelity, addressing the fundamental question of whether the model can produce valid and realistic covariance matrices, and with a comparison against a standard VAE approach. Our final analysis quantifies the impact of this data on cross-subject classification performance to determine the practical value of the proposed method.

\mysubsection{Latent Space Structure} UMAP \cite{mcinnes_2018_umap} visualization (Fig. \ref{fig:umap_space}) reveals that latent codes organize into a unified structure where subjects are heavily intermingled rather than clustered by individual. This suggests the model learned a largely subject-invariant representation---a critical property enabled by parallel transport alignment---implying generated samples will reflect generalized task patterns rather than subject-specific details.

\mysubsection{Fidelity Assessment}
Across all folds, 100\% of synthetic matrices from both prior and posterior generators passed symmetry and positive-definiteness verification checks, confirming the effectiveness of the architecture's geometric constraints and numerical stabilisation steps.%

Tab. \ref{tab:variance_analysis} compares the statistical variance and geometric spread of synthetic data relative to the original. The chosen $\gamma$ maintained statistical variance close to the original. To address the lower geometric diversity of synthetic samples, the noise vector is scaled by $\epsilon_i = 2.2$ during generation, increasing the mean intra-class Riemannian distance to $\approx 1.95$, closely matching the original data's spread ($2.03$) without distorting statistical properties. 

\mysubsection{Cross-Subject Classification Performance}
The impact of data augmentation was evaluated by comparing classification accuracies under different augmentation conditions using Wilcoxon signed-rank tests with Bonferroni correction ($p < 0.0083$). As detailed in Tab. \ref{tab:main_results}, data augmentation produced divergent effects. For the KNN classifier, augmentation consistently and significantly improved performance. Posterior-based synthetic-only training yielded the largest gain (+3.49\%, $p=0.002$), while augmented training provided +2.45\% ($p=0.002$). Prior generation produced similar but slightly smaller significant benefits (+3.00\% synthetic-only, $p<0.001$; +2.19\% augmented, $p=0.003$). In contrast, SVC performance significantly degraded with augmentation (up to -4.01\%, $p=0.002$), while MDM remained largely unaffected. %
Figs. \ref{fig:improvement_dist_prior} and \ref{fig:improvement_dist_posterior} illustrate subject-wise distributions, revealing high variability: KNN augmentation yielded gains up to +7.8\% for subject no.3 (prior generation, synthetic only condition).
A scrambled label test confirmed classifiers learned meaningful features, yielding chance-level accuracy ($\approx 50\%$) on randomized data.

\mysubsection{Standard VAE Comparison}
To validate the Riemannian framework, we compared the proposed RGP-VAE against a standard Euclidean VAE. The standard VAE failed to generate valid data, with $>40\%$ of outputs in every fold violating positive-definiteness. Furthermore, augmenting with the valid portion of its data significantly degraded MDM performance ($-9.49\%$, $p < 0.001$) and offered no statistically significant benefit to KNN or SVC. This confirms that the proposed architecture's geometric constraints are essential for generating valid and useful SPD matrices in this domain.

\mysection{discussion}

This study investigated whether the proposed RGP-VAE could generate high-fidelity EEG covariance matrices to improve cross-subject MI-BCI classification. 

\mysubsection{Generative Fidelity and Validity}
A primary contribution of this work is confirming that the RGP-VAE framework inherently generates valid SPD matrices---a non-trivial task where standard Euclidean VAEs failed (producing $>40\%$ invalid matrices). This success is attributable to the
underlying Riemannian geometry that enforces the
SPD constraint by design. Parallel Transport enabled the model to learn a subject-invariant latent space, a critical property for cross-subject generalization.

While valid, the synthetic data exhibited a slightly elevated statistical variance (ratio $\approx 1.06$) but reduced geometric diversity (ratio $\approx 0.95$). With the chosen parameters ($\gamma = 0.035$, $\epsilon_i = 2.2$), the model generated more prototypical samples concentrated near the class geometric means rather than spanning the full outlier range of real-world data.

\mysubsection{Classifier-Dependent Utility}
The impact of the synthetic data was highly divergent, revealing that data augmentation utility is not universal but classifier-dependent. 
Augmentation yielded statistically significant improvements for the KNN classifier, with posterior sampling boosting performance up to $+3.49\%$ ($p=0.002$). 
KNN likely benefits because the prototypical synthetic samples densify the class manifolds, creating more dense and reliable local neighbourhoods for distance-based classification. Conversely, performance significantly degraded for the SVC (up to $-4.01\%$, $p=0.002$). The reduced diversity of synthetic data likely caused the SVC to learn decision boundaries too narrowly fitted around class centres, reducing generalization to boundary-case real samples. 
Meanwhile, performance for the MDM classifier remained stable---a positive result compared to the standard VAE, which caused a massive degradation ($-9.49\%$ under posterior, synthetic-only condition). Unlike the naive Euclidean approach that failed to even generate valid SPD matrices, the RGP-VAE preserved the SPD validity and successfully learnt Riemannian class means. Beyond immediate classification impacts, synthesizing this data holds broader practical value; it provides a mechanism to test pipeline scalability, mitigates data scarcity---possibly for data hungry models---and enables privacy protection by avoiding raw signal sharing.

\mysubsection{Future Research Directions}
This study provides a foundational proof of concept that opens avenues for future research. Building on these findings, future work may explore advanced manifold sampling techniques, such as Riemannian Hamiltonian VAEs or Riemannian Monte Carlo sampling, to capture complex latent distributions more faithfully \cite{chadebec2021data, ChadebecRHMC}. Additionally, as demonstrated by vEEGNet \cite{vEEGNet}, integrating the RGP-VAE's geometric constraints and subject-invariance with discriminative frameworks could potentially yield latent spaces that are simultaneously geometrically valid, subject-invariant and class-discriminative. %

\mysection{conclusion}\label{sec:conclusion}

This paper developed and validated a novel Riemannian Geometry-Preserving
VAE (RGP-VAE) for generating synthetic EEG covariance matrices in the challenging cross-subject MI-BCI context. The RGP-VAE is not only capable
of consistently generating valid SPD matrices---overcoming the limitations of standard VAEs---but also closely matches the original data diversity. The high-fidelity synthetic data can maintain or even significantly improve classification performance for specific classifiers. However, divergent classifier results highlight generative capabilities on the SPD manifold do not guarantee universal downstream improvements.

\mysection{references}

\printbibliography[heading=none]

\end{document}